\documentclass[11pt, a4paper, copyright, gr]{google}

\usepackage[authoryear, sort&compress, round]{natbib}
\usepackage{wrapfig}
\usepackage{tcolorbox}

\definecolor{figGreen}{HTML}{2E7D32}
\definecolor{figRed}{HTML}{C62828}

\uselogo{}

\title{Hallucinations Undermine Trust; Metacognition is a Way Forward}

\correspondingauthor{gal.yona@google.com}

\reportnumber{} 

\renewcommand{\today}{}

\author[1]{Gal Yona}
\author[2]{Mor Geva}
\author[1]{Yossi Matias}

\affil[1]{\thepa{}{}}
\affil[2]{Tel Aviv University}

\begin{abstract} Despite significant strides in factual reliability, errors---often termed hallucinations---\textbf{remain a major concern for generative AI}, especially as LLMs are increasingly expected to be helpful in more complex or nuanced setups. Yet even in the simplest setting—factoid question-answering with clear ground truth --- frontier models without external tools continue to hallucinate.  We argue that \textbf{most factuality gains in this domain have come from \emph{expanding the model's knowledge boundary}} (encoding more facts) \textbf{rather than \emph{improving awareness of that boundary}} (distinguishing known from unknown). We conjecture that the latter is inherently difficult: models may lack the discriminative power to perfectly separate truths from errors, creating an unavoidable tradeoff between eliminating hallucinations and preserving utility. 

\par\vspace{1em}

\textbf{This tradeoff dissolves under a different framing.} If we understand hallucinations as \textbf{\emph{confident} errors}---incorrect information delivered without appropriate qualification---\textbf{a third path emerges beyond the answer-or-abstain dichotomy: expressing uncertainty.} We propose \textbf{\emph{faithful uncertainty}}: aligning linguistic uncertainty with intrinsic uncertainty. This is one facet of \textbf{\emph{metacognition}}---the ability to be aware of one's own uncertainty and to act on it. For direct interaction, acting on uncertainty means communicating it honestly; for agentic systems, it becomes the control layer governing when to search and what to trust. Metacognition is thus essential for LLMs to be both trustworthy and capable; we conclude by highlighting open problems for progress towards this objective. \end{abstract}

\begin{document}

\maketitle


\vfill

\begin{figure*}[h!]
    \centering
    \includegraphics[width=0.98\linewidth]{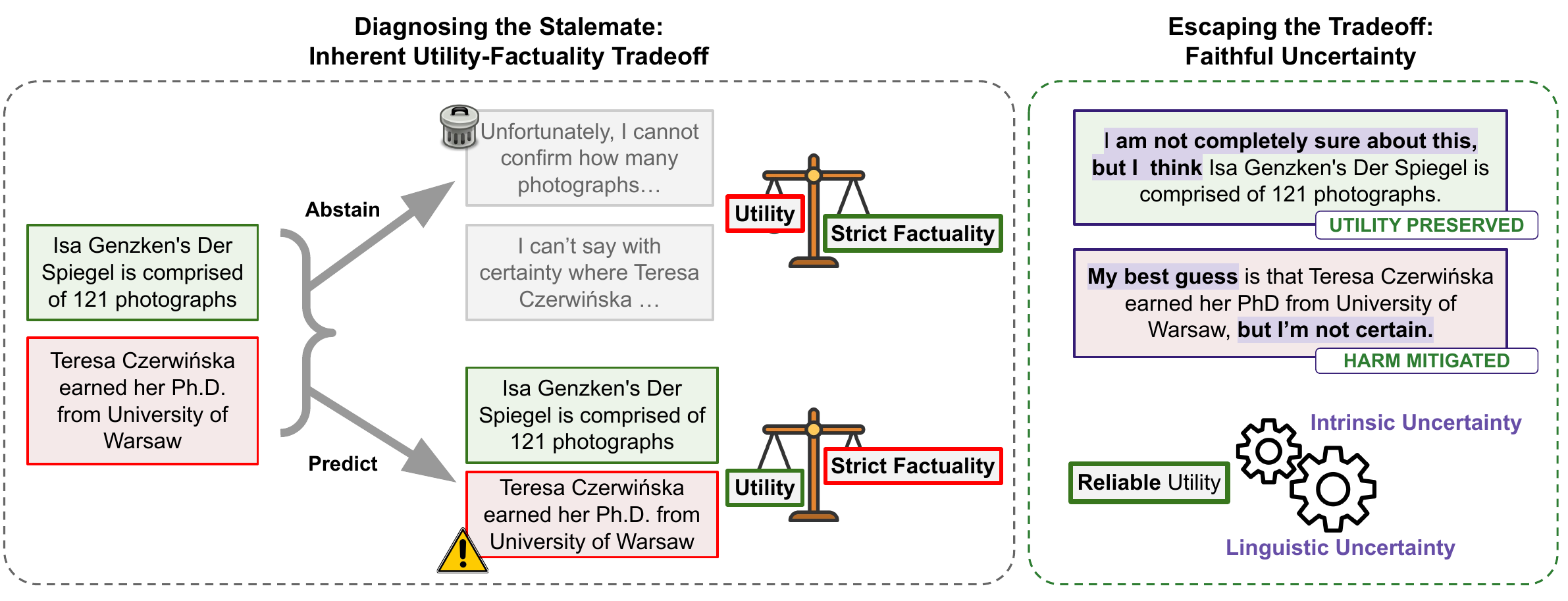}
    \captionsetup{margin=1.5cm} 
    \caption{\textbf{Escaping the Utility-Factuality Trade-off.}
    \textbf{(Left)} We conjecture that models lack the discriminative power to perfectly separate truths (\textcolor{figGreen}{green}) from errors (\textcolor{figRed}{red}). Under the traditional view that any error constitutes a hallucination, this creates a dilemma: the model must either \emph{abstain} (top), paying a ``utility tax'' by suppressing valid information (\includegraphics[height=0.9em]{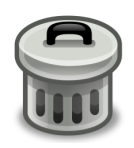}); or \emph{predict} (bottom), risking confident errors that erode trust (\includegraphics[height=0.9em]{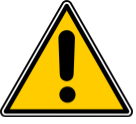}).
    \textbf{(Right)} We propose reframing hallucination as \emph{confident} error, which reveals a third path: \emph{faithful expression of uncertainty}. By aligning linguistic output with intrinsic confidence, the model retains valid information as appropriately hedged estimates while rendering errors less harmful.}
    \label{fig:main_concept}
\end{figure*}

\vfill

\clearpage

\section{Introduction}

Despite significant strides in factual reliability \citep{tian2023fine, wei2024long, grattafiori2024llama, cheng2025facts}, errors---often termed ``hallucinations''---remain a major concern for generative AI, especially as large language models (LLMs) are increasingly expected to be helpful in more complex or nuanced setups. These factually incorrect generations are often delivered with an authoritative tone, risking undermining user
trust and spreading misinformation \citep{ji2023survey, zhang2025siren, steyvers2025large}.

In this paper, we focus on a simple setting where frontier models still hallucinate: factoid question-answering with clear ground truth (setting aside long-form generation and cases of genuine ambiguity or contested claims). For models without access to external tools, we argue that \textbf{most factuality gains in this domain have come from \emph{expanding the model's knowledge boundary} (encoding more facts) rather than \emph{improving awareness of that boundary} (distinguishing known from unknown)}. We conjecture that this asymmetry arises because while expanding the knowledge boundary is often achievable through scale, data, and improved training recipes, \textbf{models may fundamentally lack the discriminative power to perfectly separate truths from errors}.

Although research on uncertainty quantification has shown that well-calibrated confidence signals can be extracted from modern LLMs \citep{kadavath2022language, lin2022teaching, tian2023just, nakkiran2025trained}, \emph{calibration} (confidence scores matching the probability of correctness) does not guarantee \emph{discrimination} (confidence scores that can sharply distinguish correct from incorrect answers). \textbf{Under the traditional view that treats hallucination as synonymous with error, limited discrimination creates an inherent tradeoff between eliminating hallucinations and preserving utility:} to guarantee zero hallucinations, a model must abstain whenever uncertain, suppressing valid information along with errors. \textbf{In practice, model providers are often unwilling to pay this ``utility tax,'' resulting in models that prioritize answering and still hallucinate} (Figure~\ref{fig:main_concept}, left).

This perspective ties together many recent empirical observations. The poor generalization of truthfulness probes \citep{levinstein2023still, orgad2024llms, sky2024androids, marks2023geometry} and the existence of confident hallucinations \citep{simhi2025hack, wang2025bias, taubenfeld2025confidence} both demonstrate the limits of model internals for veracity prediction. The failure of advanced alignment techniques, such as training models to ``confess'' errors \citep{joglekar2025training}, to mitigate hallucinations confirms that such deficits persist under strong supervision. Together, these provide evidence for the existence of a discrimination gap. Finally, the surprising finding that extended reasoning \emph{increases} hallucinations \citep{jaech2024openai, yao2025reasoning, li2025hallucination} and degrades abstention \citep{kirichenko2025abstentionbench, zhang2025factguard} reflects how current training incentivizes models to favor utility in the presence of these tradeoffs.

\textbf{However, the apparent tradeoff between trust and utility dissolves under a different framing.} If we understand hallucinations not as any error, but as \textbf{confident errors}---incorrect information delivered without appropriate qualification---\textbf{a third path emerges: expressing uncertainty.} An error communicated with appropriate hedging is not a hallucination; it is a hypothesis offered for consideration. The model need not choose between utility and trust; it can preserve both by honestly communicating its uncertainty.

Naturally, simply hedging more is not the answer. A model that hedges uniformly provides no signal; the hedge can technically be calibrated to error rates while being completely uninformative at the instance level. \textbf{What is needed is \emph{faithful} uncertainty: hedging that reflects the model's actual internal state for each specific answer.} This builds on the notion proposed by \citet{yona2024can} and later \citet{ghafouri2024epistemic}, which requires aligning the model's \emph{linguistic uncertainty} (what it says) with its \emph{intrinsic uncertainty} (what it ``believes''). Faithful uncertainty is one facet of what we call \emph{metacognition}---the ability to be aware of one's own uncertainty and to act on that awareness. In direct interaction, acting on uncertainty means communicating it honestly; as we discuss later, for agentic systems it means using uncertainty to guide tool use.

\textbf{Faithful uncertainty ensures the model's response provides an honest signal of its internal state, with clear behavioral semantics}: ``I am confident'' means the model would likely give the same answer if asked again; ``I am uncertain,'' suggests it is likely to give a conflicting answer.  This is information users can act on, regardless of whether the model is ultimately correct. Crucially, \textbf{faithful uncertainty is feasible in principle}: it depends only on the model's internal states, not on solving the difficult problem of knowing when those states correspond to truth. A model may not perfectly know when it is wrong, but it \emph{can} know when it is uncertain.

\textbf{This framing acknowledges that trust can be built on imperfect knowledge, provided uncertainty is communicated honestly}---just as we trust a doctor not for omniscience, but for reliably distinguishing diagnoses from hypotheses. It also becomes increasingly urgent as models grow more capable: \textbf{as outputs become more sophisticated, they become harder for users to verify independently, making honest communication of uncertainty a safety requirement}.

While faithful uncertainty addresses cases where the model is intrinsically
uncertain, \textbf{the remaining errors---where the model is genuinely confident but wrong---are ``honest mistakes,''} addressable only through continued knowledge expansion. This highlights how the two efforts are complementary: knowledge expansion pushes the knowledge boundary further; faithful uncertainty communicates whatever boundary remains.

Metacognition has a second facet beyond expression: for agentic systems, it becomes the control layer. The shift to agentic AI systems \textbf{effectively expands the knowledge boundary}---the model can retrieve information it does not have encoded. On the surface, this might suggest that awareness of uncertainty becomes redundant: \emph{why know what you don't know if you can simply look everything up?} \textbf{But awareness of uncertainty is precisely what enables effective tool use.} Without it, a model cannot determine when to invoke a tool (leading to inefficient overuse or dangerous underuse), nor can it appropriately weigh retrieved information against its own beliefs when conflicts arise. Faithful uncertainty is thus not circumvented by tools, but rather becomes the control layer that governs them \citep{rabanser2026towards}. Yet modern search agents lack such awareness, leading to inefficient tool overuse \citep{qian2025smart, lin2025adasearch}. Instilling metacognition thus addresses not only parametric reliability, but provides the foundation for robust agentic behavior.

While LLMs are currently poor at faithfully conveying their uncertainty, we believe this problem offers tangible headroom. Recent work demonstrates promising directions through metacognitive prompting \citep{liu2025metafaith}, fine-tuning \citep{eikema2025teaching}, and model internals \citep{ji2025calibrating}. Encouraging results also include the observation that reasoning models better express their confidence \citep{yoon2025reasoning, podolak2025read} and the success of intrinsic signals as rewards in reinforcement learning \citep{prabhudesai2025maximizing, wang2025icpo, li2025confidence}.

\begin{tcolorbox}[title=Summary of Contributions, colback=blue!5, colframe=blue!75!black, sharp corners, boxsep=3pt]


\begin{itemize}[itemsep=0.33cm]
    \item We argue that most factuality gains to date in factual question-answering with clear ground truth have come from expanding the model's knowledge boundary rather than improving awareness of that boundary, and conjecture that the latter may be fundamentally difficult due to limited discriminative power.

    \item We propose reframing hallucination as \emph{confident} error rather than mere error. This reveals a third path beyond the answer-or-abstain dichotomy: expressing uncertainty. Faithful uncertainty---the metacognitive capability of aligning linguistic uncertainty with intrinsic uncertainty---directly mitigates hallucinations (as reframed) while preserving utility (Figure~\ref{fig:main_concept}, right). Unlike calibration, which is an aggregate property, faithfulness provides an instance-level guarantee: each hedge reflects that specific answer's internal state. This approach complements continued efforts in knowledge expansion, which addresses the remaining ``honest mistakes'' where models are confident but wrong.

    \item We introduce concrete recommendations for evaluating hallucination mitigation techniques, including prioritizing discriminative measures over calibration and holistically quantifying the utility cost of interventions. For researchers interested in faithful uncertainty and metacognition, we sketch key open problems as entry points.
\end{itemize}
\end{tcolorbox}

\paragraph{Organization.} We begin with background on definitions, metrics, and mitigation strategies (\S\ref{sec:background}), then present our analysis of the challenges facing strict factuality (\S\ref{sec:limits}). Next, we propose the objective of faithful uncertainty (\S\ref{sec:faithfulness}), arguing for its feasibility and favorable utility-reliability properties, and discuss its role in agentic systems (\S\ref{sec:tools}). Finally, we outline concrete recommendations for researchers (\S\ref{sec:action}), address alternative viewpoints (\S\ref{sec:alternative}), and conclude (\S\ref{sec:discuss}).

\section{Background}
\label{sec:background}

\subsection{Problem Scope}

\textbf{Extrinsic Hallucinations in Parametric Models.}
We distinguish between two modes of deployment. Parametric LLMs rely on their own parameters \citep{petroni2019language,roberts2020much}, while Tool-Augmented LLMs interact with external sources, such as search engines or APIs, to retrieve information at inference time \citep{ lewis2020retrieval, nakano2021webgpt, yao2022react, schick2023toolformer}. We primarily focus on the former and discuss implications for tool-use in \S\ref{sec:tools}. Within this scope, we specifically target extrinsic hallucinations -- generations that are factually incorrect with respect to real-world knowledge \citep{ji2023survey, huang2025survey} -- as opposed to intrinsic hallucinations (contradicting source text) or  reasoning errors.

\textbf{The Challenge of Tail Knowledge.}
Standard hallucination mitigation evaluations often center on common misconceptions \citep{lin2021truthfulqa} or head knowledge \citep{ joshi2017triviaqa, kwiatkowski2019natural}, potentially masking the severity of hallucinations in the regime of sparse data \citep{kandpal2023large, mallen2023not}. To 
probe the true boundaries of reliability, we focus on tasks that require ``long tail'' knowledge, using benchmarks \citep{wei2024measuring, haas2025simpleqa, mallen2023not, jackson2025aaomniscienceevaluatingcrossdomainknowledge} that ask explicit and simple questions about facts regarding very rare entities  (e.g., \emph{``On which U.S. TV station did the Canadian reality series To Serve and Protect debut?''}).  

\subsection{Measuring Reliability}

\textbf{The Utility-Factuality Trade-off.}
Evaluating hallucinations requires nuance, because a model can trivially achieve zero hallucinations by refusing to answer any question where it is not certain. While achieving zero hallucinations, this strategy renders the model practically useless. Robust evaluation must therefore track both \textit{accuracy} (overall correctness) and \textit{attempted accuracy} (correctness on the subset for which an answer was attempted). Since in practice the knowledge captured in the model is bounded, the ideal behavior is to maximize both accuracy and attempted accuracy, using summary metrics like F1 \citep{wei2024measuring} or Omniscience Index \citep{jackson2025aaomniscienceevaluatingcrossdomainknowledge}.

\textbf{Calibration vs. Discrimination.}
\label{sec:background:confidence}
Central to our argument is the distinction between two dimensions of uncertainty quantification. \emph{Calibration} measures the alignment between confidence scores and empirical accuracy; scores are perfectly calibrated if, among all predictions assigned confidence $p$, exactly $p\%$ are correct. In contrast, \emph{discrimination} measures the ability to distinguish correct from incorrect answers based on confidence.\footnote{A similar distinction has been drawn in the metacognition literature in cognitive science, where the analogous concept, \emph{resolution}, has been argued to be more diagnostic of metacognitive accuracy than calibration \citep{nelson1984comparison, fleming2014measure}.} Crucially, calibration does not imply discrimination. A confidence score that assigns a static confidence of $0.6$ to every answer (and is correct 60\% of the time) is perfectly calibrated yet has zero discriminative power. As we argue in \S\ref{sec:limits},  eliminating hallucinations in practice requires good discrimination, not just calibration. 

\subsection{Existing Mitigation Strategies}

\label{sec:background:mitigation}
Research on mitigating  hallucinations in parametric LLMs has generally followed two streams. Training-time interventions include careful data filtering and regularization \citep{gekhman2024does, kaplan2026fine}, penalizing non-factual outputs \citep{ouyang2022training, tian2023fine}, using  rewards  based on collaborative games \citep{eisenstein2025don} and mitigating overconfidence via linguistic calibration \citep{mielke2022reducing,  yang2024alignment, stengel2024lacie}.
Inference-time interventions focus on steering the model toward factual generations without altering weights, using custom decoding strategies \citep{chuangdola, shi2024trusting}, relying on internal signals \citep{li2023inference, yu2024mechanistic} or self-verification \citep{cohen2023lm, dhuliawala2024chain}. 


\section{Why Hallucinations Persist}
\label{sec:limits}

Having established the background and metrics, we now turn to analyze the state of the field.  In this section, we argue that the objective of fully eliminating hallucinations faces fundamental challenges. We ground this position in theoretical limits, information bottlenecks, and empirical evidence from state-of-the-art LLMs.

\begin{figure*}[ht]
    \centering
    \includegraphics[width=0.86\textwidth]{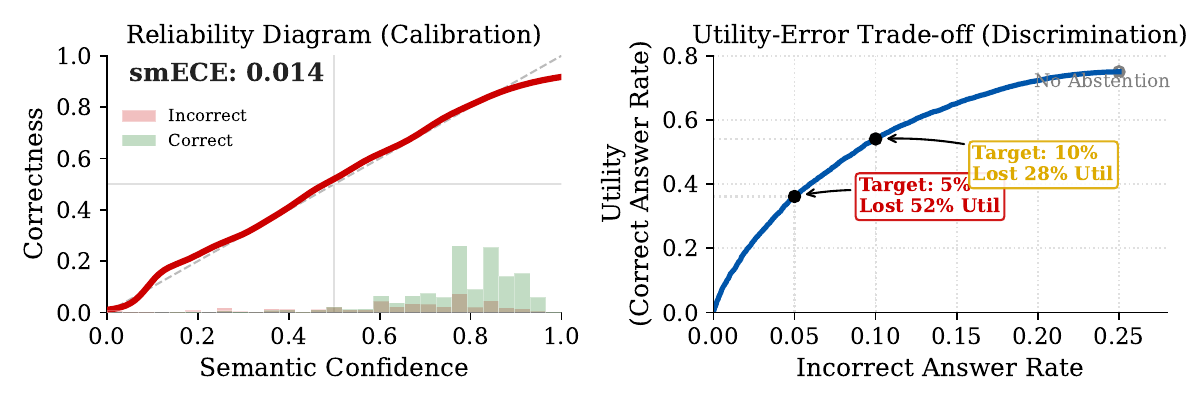}
\caption{\textbf{Calibration vs Discrimination}. \textbf{Left}: We simulate data (see App. \ref{sec:appendix}) to match the reliability diagram in \citet{nakkiran2025trained} (Figure 3). The model has a 25\% base error rate and achieves strong calibration \citep[measured w. SmoothECE;][]{blasioksmooth}, indicated by the curve hugging the diagonal, though the underlying histogram shows that correct and incorrect answers often share similar confidence scores. \textbf{Right}: The Utility-Error Trade-off curve illustrates the cost of fully eliminating hallucinations. Despite good calibration, reducing the hallucination rate from 25\% (No Abstention) to a strict target of 5\% requires discarding 52\% of valid answers (Utility). This visualizes the utility tax: without very strong discrimination, eliminating hallucinations requires suppressing a massive volume of correct information.}
    \label{fig:calibration_discrimination}
\end{figure*}

\subsection{The Theoretical Ceiling}
Previous work argued that extrinsic hallucinations are a structural inevitability of auto-regressive text generation.  \citet{banerjee2025llms} and \citet{xu2024hallucination} utilized the Halting Problem and diagonalization arguments to prove that no computable model can universally verify truth or learn all ground-truth functions. \citet{kalai2024calibrated} showed calibrated models are mathematically bound to hallucinate
when generating facts whose truth value cannot be inferred from other facts,
while \citet{kalavasis2025limits} established a formal trade-off between consistency and breadth—proving that reducing hallucination rates below a critical threshold necessitates a drastic reduction in output diversity, inevitably forcing the model into mode-collapse.

\subsection{The Discriminative Gap}
\label{sec:limits:gap}
Previous work has largely focused on the positive results in eliciting well-calibrated confidence signals from LLMs \citep{kadavath2022language, tian2023just, zhao2024fact, nakkiran2025trained}. While this demonstrates that LLMs can accurately aggregate uncertainty, there is a critical distinction between knowing the average error rate (calibration) and knowing which specific instances are errors (discrimination). We conjecture that the practical failure of standard mitigation techniques stems precisely from this deficit—a fundamental lack of discriminative power.  Indeed, a weak correlation between these two metrics has 
been demonstrated in practice for many confidence scores \citep{savage2025large, taubenfeld2025confidence, tao2025revisiting}. We further illustrate this point in \autoref{fig:calibration_discrimination}, showing a practical scenario in which using a well-calibrated confidence signal (left) to eliminate hallucinations results in major trade-offs with utility (right). E.g., to reduce the error rate from 25\% to a target of 5\%, the model must discard over 50\% of its valid answers. 

To empirically quantify the discrimination gap, we review AUROC values from the
literature for the task of separating correct from incorrect answers using a
model's confidence signal (AUROC\,=\,1.0 is perfect; 0.5 is random). Across methods, models, and tasks, AUROC
clusters in the 0.70--0.85 range for realistic factual QA tasks
\citep{farquhar2024detecting, savage2025large, kang2025uncertainty}.
Concretely, \citet{farquhar2024detecting} report an average AUROC of
0.79 across 30 model$\times$task combinations using semantic
entropy; \citet{savage2025large} find GPT-4 tops out at 0.79 in
medical QA; and \citet{kang2025uncertainty} find GPT-4o-mini reaches only
0.68--0.72 on biography generation, which resembles our tail facts setting. Crucially, this range is insufficient to escape the utility tax.
The simulation underlying Figure~\ref{fig:calibration_discrimination} has
AUROC\,=\,0.71, consistent with the literature average; at this level,
reducing the error rate from 25\% to a 5\% target requires discarding 52\% of
valid answers. Even at the 0.85 ceiling, the tax remains $\sim$28\%. The tax
only becomes negligible ($<$5\%) at AUROC\,$\geq$\,0.95, which is far above any
method currently reported for knowledge-intensive tasks.

\subsection{Corroborating Anomalies}

Recent anomalies in model development corroborate these theoretical constraints, mapping directly to the conjectured discrimination gap. First, the poor generalization of truthfulness probes \citep{levinstein2023still, orgad2024llms, sky2024androids, marks2023geometry}
and the demonstrated existence of ``confident hallucinations'' -- factual errors with high intrinsic confidence \citep{simhi2025hack, wang2025bias, taubenfeld2025confidence} -- demonstrate that in practice, the information required to robustly distinguish correct from incorrect answers is often absent even from the model's latent states.
This also explains the failure of advanced supervision in practically mitigating hallucinations. \citet{joglekar2025training} demonstrate that while models can be aligned to ``confess'' to intentional safety violations, this capability fails to transfer to hallucinations. This divergence indicates that unlike safety issues, hallucinations are not merely behavioral bugs but stem from the discrimination gap; the model cannot be aligned to report errors it cannot internally represent.
Finally, in the absence of reliable discrimination, optimizing for utility actively exacerbates hallucination. Recent work indicates that ``thinking'' 
often increases hallucination rates \citep{jaech2024openai, yao2025reasoning, li2025hallucination} and degrades abstention \citep{kirichenko2025abstentionbench}, with increased performance gaps between answerable and unanswerable questions \citep{ zhang2025factguard}. By incentivizing extended chain-of-thought and persistence, these models essentially prioritize the completion of a reasoning path over abstention, effectively rationalizing incorrect answers to satisfy the utility objective.

\begin{figure}[t]
\setlength{\belowcaptionskip}{-10pt}
    \centering
\includegraphics[width=0.8\linewidth]{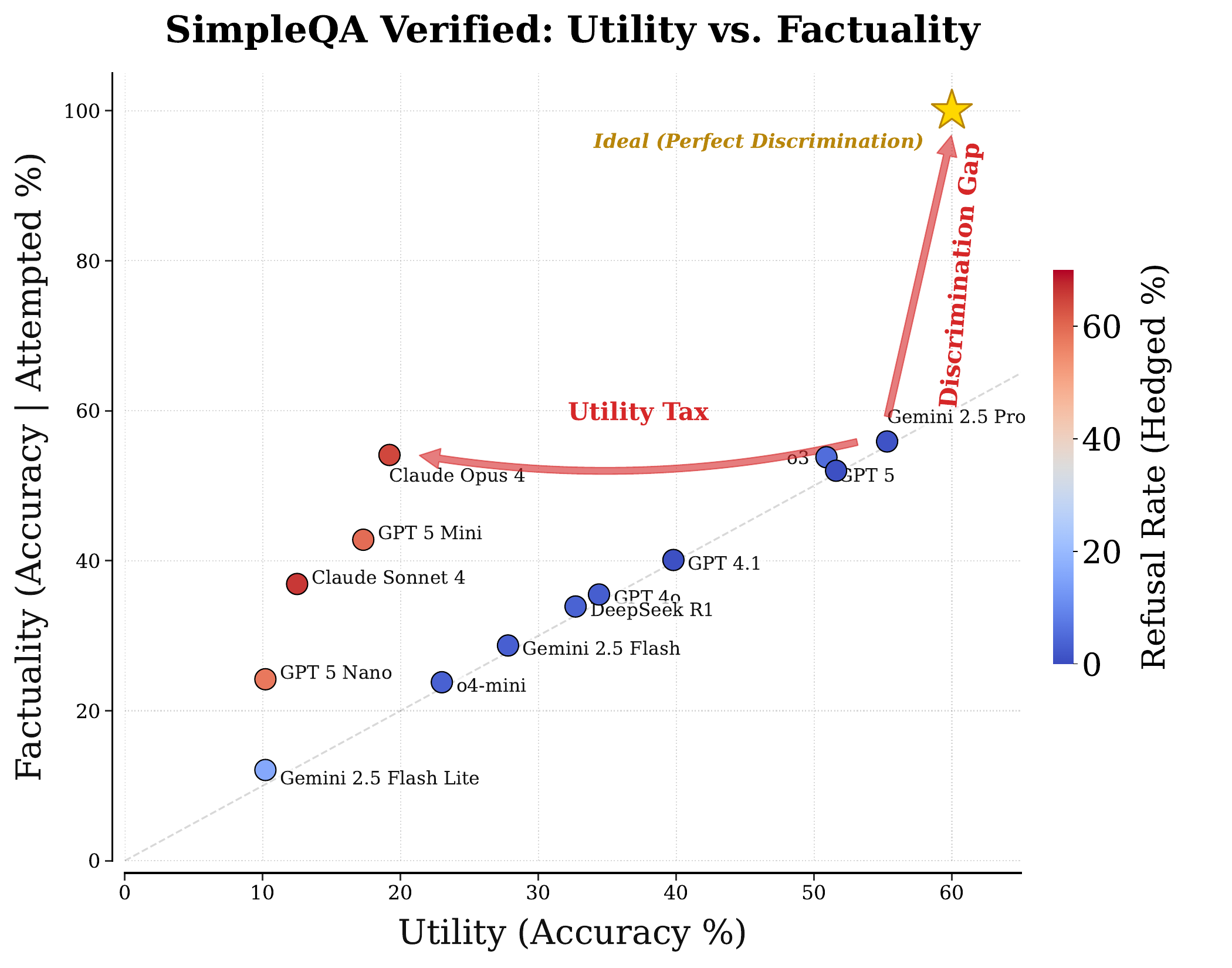}
\caption{
\textbf{The empirical tradeoff}: A visualization of results from Table 7 in SimpleQA Verified \citep{haas2025simpleqa}. Color indicates refusal rate. Most frontier models 
hug the diagonal (low refusal), optimizing for coverage at the cost of high hallucination rates. Achieving higher factuality 
requires aggressive abstention, imposing a severe utility tax (moving left). The unpopulated region near the ideal (gold star) illustrates the discriminative gap: current models lack the internal separability required to maximize factuality without destroying utility.
    }
    \label{fig:standstill}
\end{figure}

\subsection{The Empirical Tradeoff}
Lastly, we visualize the culmination of these factors in \autoref{fig:standstill}, which plots the performance of various state-of-the-art models on the SimpleQA Verified dataset from \citet{haas2025simpleqa}. The plot reveals that the field has effectively bifurcated. Most frontier models (blue circles) hug the diagonal, optimizing for coverage at the cost of high hallucination rates. Conversely, models that attempt to maximize factuality (red circles) are forced to move left rather than up, paying the utility tax by discarding valid answers. The region of ``ideal'' performance—the top-right corner—remains entirely unpopulated. This empty space visualizes the discriminative gap: it is the region that we conjecture LLMs cannot currently reach, because they lack the intrinsic capability to distinguish their own hallucinations from their knowledge.

\section{Faithful Uncertainty}
\label{sec:faithfulness}

In light of the challenges described in \S\ref{sec:limits}, we propose a pragmatic adjustment to the prevailing research objective. Rather than focusing solely on expanding knowledge, models should be as knowledgeable as possible \emph{while faithfully expressing whatever uncertainty remains}. This does not abandon the factuality objective, as knowledge expansion remains valuable where headroom exists. However, it complements it with an objective that addresses those remaining cases where knowledge falls short and reliability is difficult due to the discrimination gap.

\subsection{Defining the Objective}
We adopt the framework proposed by \citet{yona2024can}, defining faithful uncertainty as the alignment between the model's internal state and its verbalized output. We provide a self-contained overview of the main definitions and metrics in Appendix \ref{sec:appendix:faithfulness}. Specifically, faithful uncertainty requires an alignment between \emph{intrinsic uncertainty}, i.e., the model's statistical confidence in the semantic meaning of its assertion (where high uncertainty implies a high probability of generating conflicting answers), and 
\emph{linguistic uncertainty}, defined as the confidence expressed with words in the model's generated response, e.g., using phrases like ``I am 90\% sure,'' or ``I might be mistaken''. A model is said to \emph{faithfully express its uncertainty} if its linguistic uncertainty accurately reflects its intrinsic uncertainty. Unlike fully eliminating hallucinations, which requires the model's output to match the \emph{external world}, faithful uncertainty requires the model’s output to match its \emph{internal state}.

\subsection{The Feasibility Argument}
Faithful uncertainty bypasses the limitations discussed in \S\ref{sec:limits}. While mapping finite parameters to an infinite world is theoretically limited \citep{xu2024hallucination}, mapping internal parameters to an output string is a fully observable, closed-loop problem.
Even if there is no universal ``truth direction'' in the activation space that allows for perfect discrimination, the confidence signal is inherently computable from the model's weights. The model does not need to know if the probability $P(\text{answer})=0.6$
 corresponds to ``truth'' in the real world; it only needs to detect that its internal confidence is $0.6$
 and map that signal to a verbalized hedge. Because the ground truth for faithfulness is internal to the system, it is theoretically solvable through architectural improvements, data modifications  and better training recipes.

 \subsection{Reliable Utility}
The faithful uncertainty objective directly addresses the utility tax (\S\ref{sec:limits:gap}).  Consider a set of answers where the model has 60\% intrinsic confidence. If the confidence is well-calibrated, exactly 60\% of these answers are correct. Under the goal of fully eliminating hallucinations, the model must make a collective decision for this entire set: to avoid the 40\% of hallucinations, it must abstain on the whole cluster, thereby discarding the 60\% of correct answers and hindering utility. Conversely, under the faithful uncertainty paradigm, the model preserves this utility: It generates the answers, but wraps them in appropriate epistemic markers. In this framework, a confident error remains an hallucination (albeit a faithful one), but an error wrapped in appropriate uncertainty is transformed into a useful hypothesis. 

We define this outcome as \emph{reliable utility}: the ability to maximize the volume of provided information without compromising user trust, achieved by aligning the decisiveness in which a claim is conveyed to the model's intrinsic confidence in it. Reliable utility mimics the way trust is established in human professionals. For example, we value a doctor not because they are all-knowing (omniscient), but because they faithfully communicate the distinction between a diagnosis they are certain of and a hypothesis they are merely testing. By allowing the model to answer when confident and hedge when uncertain, we can make models more trustworthy \citep{kim2024m, zhou2024relying} without making them less practically useful.

\subsection{The Research Opportunity: Tractable Headroom}
While this objective is theoretically feasible, it currently remains an unbridged gap. \citet{yona2024can} have demonstrated that current state-of-the-art models are far from satisfying this desideratum; they typically express high linguistic confidence even when their intrinsic uncertainty is low.
Recent work has already begun to explore this capability, showing promising results using methods ranging from meta-cognitive prompting strategies \citep{liu2025metafaith}, supervised fine-tuning \citep{eikema2025teaching}, and steering based on internal representations \citep{ji2025calibrating}. Together with continued efforts to expand model knowledge, faithful uncertainty offers a path  
  toward models that are both more knowledgeable and more trustworthy.  We detail concrete challenges for advancing this direction in \S\ref{sec:action}.

\section{Metacognition in the Age of Agents}
\label{sec:tools}

We have argued that fully eliminating hallucinations requires strong discrimination—the ability to separate what models know from what they don't—and that this is fundamentally difficult. The dominant strategy of \emph{agentic AI} \citep{yao2022react, wang2024survey} might seem to sidestep this problem: with access to external tools, a model can in principle look up any fact. \emph{Why know what you don't know if you can simply search?} We argue the opposite: tools do not remove the need for faithful uncertainty but amplify it. Without awareness of its own uncertainty, a model cannot determine when to invoke a tool (leading to inefficient overuse or dangerous under-use), nor can it appropriately weigh retrieved information against its own beliefs when conflicts arise. Faithful uncertainty thus becomes the control layer that governs tool use.

\begin{figure}[h]
\centering
\includegraphics[width=0.6\textwidth]{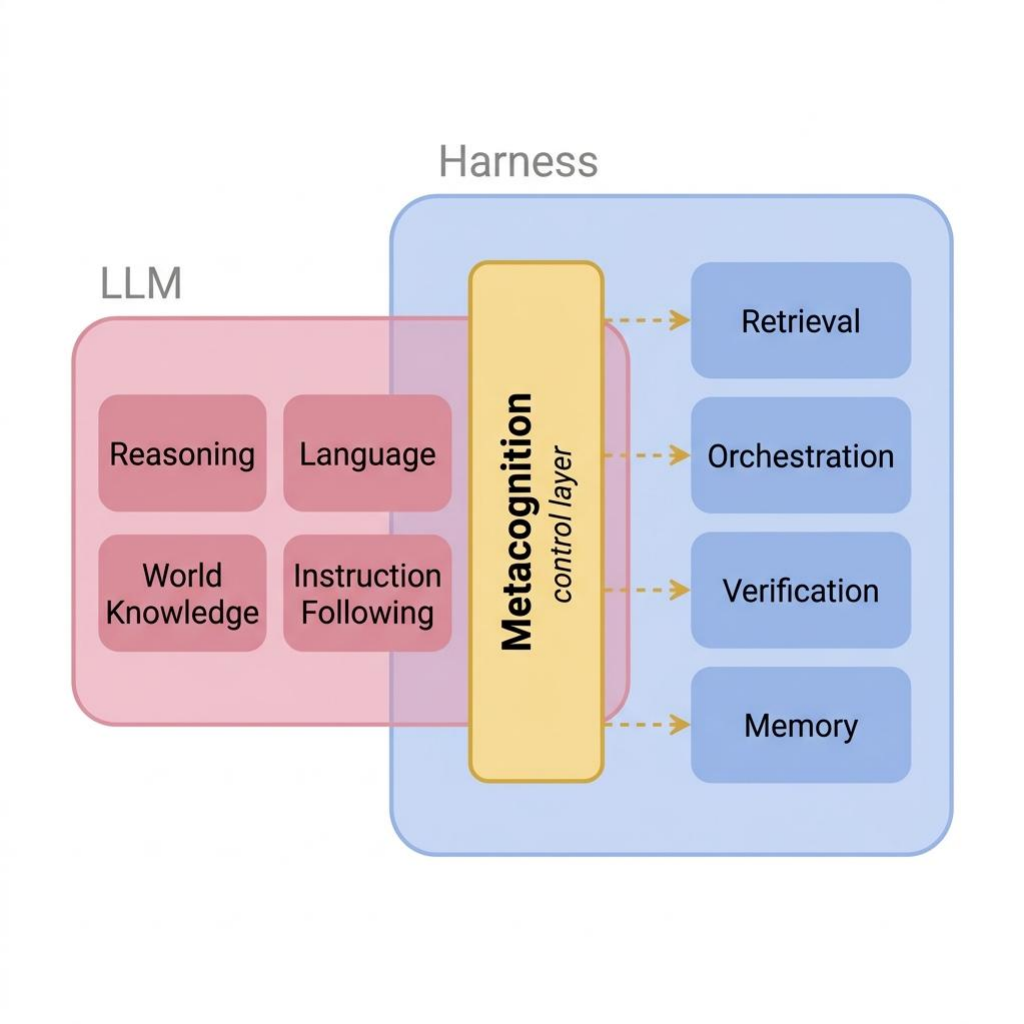}
\caption{
\textbf{A model that knows what it doesn't know can make the harness smarter and simpler.} The model's metacognition acts as the API or \emph{control layer} (yellow) between the underlying LLM (pink) and the harness (blue). Without it, the harness is ``flying blind'' – it has to make all grounding and routing decisions externally, based on query type heuristics. With it, the agent can dynamically regulate behavior, e.g. retrieve only when confidence is low (efficiency) and express doubt when retrieved evidence conflicts with internal priors (reliability). 
}
    \label{fig:agents}
\end{figure}

\textbf{Tool-Use Masks the Reliability Problem.}
Current evaluations obscure this need. By focusing on final output accuracy, benchmarks reward successful retrieval without testing whether the model understood why it needed to search. Low hallucination rates may reflect retrieval quality rather than metacognitive competence—creating systems that are factually correct but unaware of their own limitations. This fragility is exposed when tools fail or return unexpected results; indeed, recent work \citep{qian2025smart, lin2025adasearch, yan2026act, xu2026trust} shows that modern search agents lack such self-awareness, leading to systematic overuse.

\textbf{Storage vs Control.}
Tools solve what we call the storage problem: the model need not encode every fact. But they introduce a control problem: governing the process of retrieval, verification, and orchestration — functions collectively managed by the \emph{agent harness}, the scaffold that processes inputs, routes tool calls, and returns results. An agent must judge when its internal knowledge suffices and when to delegate to the harness -- a decision defined by its uncertainty. When retrieval returns conflicting or low-quality information, the agent must weigh these signals against internal priors rather than blindly accepting whatever appears in context \citep{petroni2020context, li2023large}. As illustrated in \autoref{fig:agents}, faithful uncertainty underlies all such control decisions.

\textbf{Towards Metacognitive LLMs.} Drawing on human metacognition \citep{james1890principles, son2002relation}, we emphasize two processes: \emph{introspection} (assessing one's own uncertainty) and \emph{regulation} (adjusting behavior based on that assessment). Contemporary agents often rely on static heuristics or over-engineered harnesses. Future agents in open-ended environments require dynamic control: determining when information suffices, when to verify, when to halt. 
Instilling metacognition is thus not only a complement to eliminating hallucinations, but a prerequisite for reliable autonomous agents.

\section{Call to Action}
\label{sec:action}

In this section, we offer concrete recommendations for the research community. We divide these into two categories: an overview of the main challenges and open problems for those exploring our suggested metacognitive LLMs and faithful uncertainty objectives,  and practical suggestions for research on direct hallucination mitigation.

\begin{figure}[t]
    \centering
    \includegraphics[width=0.75\textwidth]{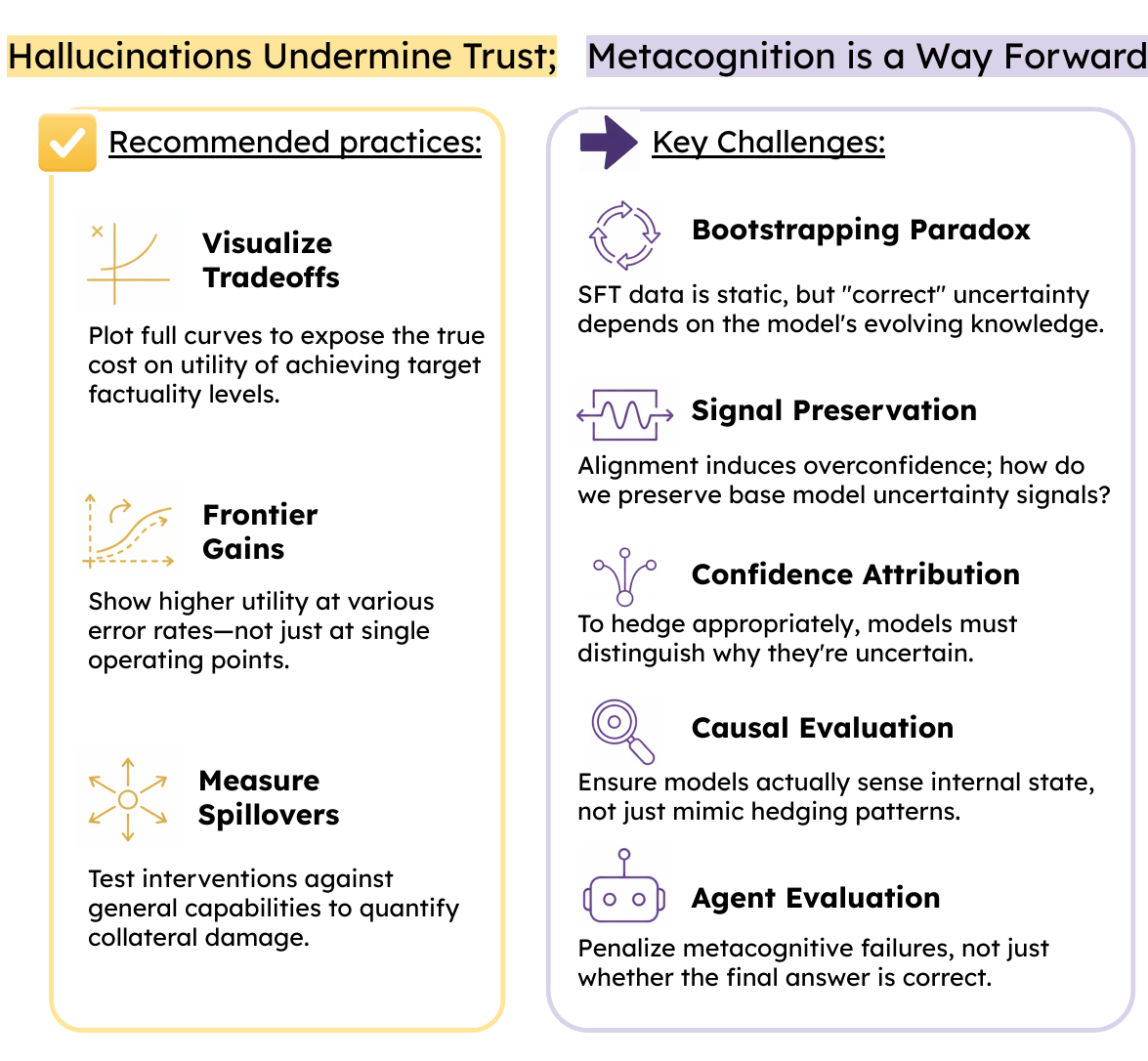}
    \caption{Recommendations for the research community.}
    \label{fig:placeholder}
\end{figure}

\subsection{Challenges for Metacognitive LLMs}

Unlocking models that can faithfully reflect their uncertainty requires solving several unique methodological hurdles:

\textbf{The Bootstrapping Paradox.}
Base models, trained on authoritative internet text, rarely express doubt naturally \citep{yona2024can, zhou2024relying}. Teaching the syntax of hedging (e.g., \emph{``I am not entirely sure...''}) therefore requires supervised fine-tuning (SFT), but this creates a paradox: SFT datasets are static, whereas the ``correct'' uncertainty label is dynamic relative to the model's current state. Training on a static label of \emph{``I don't know''} for a fact the model happens to know (or vice versa) induces hallucinated uncertainty or confidence. This requires developing  infrastructure to support such dynamic datasets or  methods to bootstrap the behavior of uncertainty (via SFT) without overfitting to possibly stale knowledge boundaries.

\textbf{Preserving the Signal through Post-Training.}
Growing evidence suggests pre-trained models possess well-calibrated uncertainty representations, that are degraded during post-training \citep{kadavath2022language, achiam2023gpt, tian2023just, zhu2023calibration}. Standard alignment techniques tend to induce mode-seeking behavior, making aligned models significantly more overconfident than their base counterparts \citep{he2025rewarding, song2025outcome}. If our goal is to be faithful to the model's true knowledge boundary (often best captured by the base model), we face a conflict: how do we align models for safety and instruction following without erasing the subtle distributional information required for instilling metacognition? Developing  such ``uncertainty preserving'' alignment algorithms is therefore an important direction for future work.

\textbf{Linguistic Precision Requires Confidence Attribution.}
To  effectively communicate uncertainty, a single scalar confidence score is insufficient \citep{delacroix2025beyond}, as a model may be uncertain for widely different reasons: ambiguity in the prompt (aleatoric), lack of knowledge (epistemic), ambiguity regarding the alignment behavior (normative), etc. High-quality faithful expressions of uncertainty therefore require the ability to trace the source of the uncertainty and map it to the appropriate linguistic hedge (e.g.,  \emph{``It depends on what you meant by X''} vs. \emph{``I do not recall X''}). Research should focus on disentangling these sources to yield a more informative and actionable signal.

\textbf{Rigorous Causal Evaluation.}
A fundamental scientific risk is that models may learn to mimic the \emph{style} of faithful uncertainty without the substance, by learning simple heuristics (e.g., ``always hedge when the prompt contains rare entities'') rather than actually sensing their internal state. This is a general challenge in evaluating metacognitive abilities in LLMs, with recent works proposing approaches such as concept injection \citep{lindsey2025emergent}, cross-model evaluations \citep{kapoor2024large, binderlooking, li2025training, song2025privileged} and strategic games in which the model benefits from assessing and utilizing its own confidence \citep{ackerman2026evidence}. Developing proper evaluation frameworks is therefore a fundamental aspect of the overall research effort.

\textbf{Evaluating Metacognition in Agents.}
Finally, as we move to agents, evaluation must shift from end-to-end correctness to process-based control. Standard benchmarks often reward agents that ``luck into'' the right answer despite bad reasoning \citep{yona2025keep}. Agent evaluations could benefit from \emph{control-centric} and model-dependent evaluations that seek to penalize metacognitive failures, such as searching for known facts (inefficiency) or trusting sources that conflict with known knowledge (sycophancy), regardless of whether the final answer happens to be correct.

\subsection{Better Hallucination Mitigation Evaluations}
\label{sec:action:hallucinations}

The key challenge we identify is that a utility tax for eliminating hallucinations is unavoidable, and current evaluation paradigms are insufficient to fully capture and reflect it. Thus, for researchers working on methods aiming to fully eliminate hallucinations, 
we recommend three concrete evaluation practices:

\textbf{Visualize the Utility-Error Trade-off.} We recommend moving away from calibration-based metrics (ECE), which mask the discriminative gap by averaging over the whole distribution, but also away from summary discrimination metrics (like AUROC), which obscure specific operating costs. As we argued in \S\ref{sec:limits}, a model can be well-calibrated or have a decent AUROC yet still require prohibitive abstention rates to achieve high reliability. Instead, researchers should visualize the full Utility-Error Curve, as shown in Figure \ref{fig:calibration_discrimination} (Right), which explicitly exposes how much utility must be sacrificed to achieve a specific target error rate.

\textbf{Demonstrate Frontier Improvements.}
Claims of ``reducing hallucinations'' often amount to simply sliding along the existing tradeoff curve (e.g. increasing the refusal threshold) rather than improving the  model's underlying capability. We urge the community to reject comparisons based on single operating points (e.g., ``we achieved 95\% accuracy''); Instead, contributions should 
demonstrate that for a fixed error rate, the method yields higher utility than the baseline.

\textbf{Measure Holistic Spillovers.} Finally, interventions must be tested for ``collateral damage.'' Tuning a model to refuse long-tail queries often causes it to become evasive on ``head'' knowledge or less helpful in reasoning coding or creative tasks. We recommend evaluating against a suite of such tasks  to fully quantify the ``cost'' of an intervention as the 
degradation of helpfulness across the model's general capabilities, not just the lost recall on the target set.

\section{Alternative Viewpoints}
\label{sec:alternative}

\subsection{We Should Not Be Deprioritizing Factuality}                                         
A skeptic might argue that by emphasizing faithful uncertainty, we risk diverting attention from the important work of making models more knowledgeable. If resources shift toward metacognition, does progress on factuality slow?

We stress that faithful uncertainty is not a replacement for knowledge expansion, but a complement to it. Model providers already have strong incentives to develop models that are as knowledgeable as possible, and these efforts should continue. Moreover, different domains offer different headroom: in emerging areas like multimodal understanding, there remains substantial room for basic factuality improvements before the discrimination gap becomes the limiting factor \cite{cheng2025facts}. Our proposal addresses cases where knowledge expansion alone falls short -- when models encounter questions at the edge of their competence, faithful uncertainty ensures they communicate limitations rather than confabulate. The two objectives are synergistic: a more knowledgeable model with good metacognition is strictly better than either capability alone.

\subsection{Users Prefer Confidence Over Uncertainty}

From a product perspective, one might argue that users prefer decisive answers, and that this is partly why RLHF-aligned models are decisive. Constant hedging 
creates friction and may be perceived as incompetence, especially in creative or high-velocity tasks like coding and writing.

We stress that this objection ignores the specific scope of our argument
(\S\ref{sec:background}): faithful uncertainty, by design, does not target creative domains in which hallucination is indeed desirable. Furthermore, in long-form generation, faithful uncertainty need not be intrusive, and ``localized'' expressions of uncertainty, such as flagging a specific line of code or a specific date, 
can add value without blocking the user. 

\subsection{Latent Truth Exists, We Just Need Better Probes}

A strong counter-position is that the challenges we describe in \S\ref{sec:limits} are overstated. Proponents could argue that LLMs are incentivized to encode representations of truth \citep{ravfogel2025emergence, marks2023geometry}, so the main bottleneck may lie with current methods not yet sophisticated enough to extract it \citep{liu2023cognitive, orgad2024llms, gekhman2025inside}. From this perspective, the pursuit of better discrimination should continue rather than be deprioritized.

We view the search for latent truth as a valuable pursuit that may well ease the tradeoffs we describe. However, it requires the strong assumption that a universal truth representation exists for the entire long tail of facts—an assumption we are skeptical of given the evidence in \S\ref{sec:limits:gap}. Faithful uncertainty, by contrast, offers concrete headroom today. Unlike the ``truth direction,'' we have robust evidence that models already possess accessible confidence signals that can be exploited. First, recent work in mechanistic interpretability demonstrates the feasibility of distilling self-awareness and confidence directly from the model \citep{stolfo2024confidence}. Second, reasoning models have been shown to be significantly better at expressing their confidence \citep{yoon2025reasoning, podolak2025read}—even while hallucinating more—suggesting that the metacognitive signal is distinct from the factual one. Finally, intrinsic signals \citep{chentanez2004intrinsically} are already used as rewards in RL to encourage diversity \citep{kayal2025impact, sukhija2025optimism} and improve reasoning \citep{prabhudesai2025maximizing, wang2025icpo, li2025confidence}.


\section{Discussion}
\label{sec:discuss}

We have argued that fully eliminating hallucinations faces fundamental challenges due to a discrimination gap, and proposed faithful uncertainty as a complementary objective. This metacognitive awareness becomes increasingly important as LLMs evolve into agentic systems, where it serves as the control layer for robust tool use.

Faithful uncertainty connects to broader objectives in AI safety: at its core, it is a form of honesty—requiring models to accurately represent their epistemic state rather than project false confidence. Crucially, uncertainty communication enables appropriate human oversight, inviting users to verify, seek additional sources, or exercise their own judgment when models express doubt. Realizing this vision requires 
a shift in both model development (as current benchmarks focus on factual accuracy) and user expectations (users that expect LLMs to express their uncertainty, and can interpret that uncertainty appropriately).

\section*{Acknowledgments}

We thank Jonathan Berant, Alain Vaucher and Nitay Calderon for providing helpful comments on earlier drafts of this work.

\bibliography{example_paper}

@article{wei2024long,
  title={Long-form factuality in large language models},
  author={Wei, Jerry and Yang, Chengrun and Song, Xinying and Lu, Yifeng and Hu, Nathan and Huang, Jie and Tran, Dustin and Peng, Daiyi and Liu, Ruibo and Huang, Da and others},
  journal={Advances in Neural Information Processing Systems},
  volume={37},
  pages={80756--80827},
  year={2024}
}

@article{farquhar2024detecting,
  title={Detecting hallucinations in large language models using semantic entropy},
  author={Farquhar, Sebastian and Kossen, Jannik and Kuhn, Lorenz and Gal, Yarin},
  journal={Nature},
  volume={630},
  number={8017},
  pages={625--630},
  year={2024},
  publisher={Nature Publishing Group UK London}
}

@article{kang2025uncertainty,
  title={Uncertainty quantification for hallucination detection in large language models: Foundations, methodology, and future directions},
  author={Kang, Sungmin and Bakman, Yavuz Faruk and Yaldiz, Duygu Nur and Buyukates, Baturalp and Avestimehr, Salman},
  journal={arXiv preprint arXiv:2510.12040},
  year={2025}
}

@article{nelson1984comparison,
  title={A comparison of current measures of the accuracy of feeling-of-knowing predictions.},
  author={Nelson, Thomas O},
  journal={Psychological bulletin},
  volume={95},
  number={1},
  pages={109},
  year={1984},
  publisher={American Psychological Association}
}

@article{fleming2014measure,
  title={How to measure metacognition},
  author={Fleming, Stephen M and Lau, Hakwan C},
  journal={Frontiers in human neuroscience},
  volume={8},
  pages={443},
  year={2014},
  publisher={Frontiers Media SA}
}

@inproceedings{
ackerman2026evidence,
title={Evidence for Limited Metacognition in {LLM}s},
author={Christopher Ackerman},
booktitle={The Fourteenth International Conference on Learning Representations},
year={2026},
url={https://openreview.net/forum?id=gb9HR8hxtU}
}

@article{song2025privileged,
  title={Privileged Self-Access Matters for Introspection in AI},
  author={Song, Siyuan and Lederman, Harvey and Hu, Jennifer and Mahowald, Kyle},
  journal={arXiv preprint arXiv:2508.14802},
  year={2025}
}

@article{grattafiori2024llama,
  title={The llama 3 herd of models},
  author={Grattafiori, Aaron and Dubey, Abhimanyu and Jauhri, Abhinav and Pandey, Abhinav and Kadian, Abhishek and Al-Dahle, Ahmad and Letman, Aiesha and Mathur, Akhil and Schelten, Alan and Vaughan, Alex and others},
  journal={arXiv preprint arXiv:2407.21783},
  year={2024}
}

@article{cheng2025facts,
  title={The FACTS Leaderboard: A Comprehensive Benchmark for Large Language Model Factuality},
  author={Cheng, Aileen and Jacovi, Alon and Globerson, Amir and Golan, Ben and Kwong, Charles and Alberti, Chris and Tao, Connie and Ben-David, Eyal and Tomar, Gaurav Singh and Haas, Lukas and others},
  journal={arXiv preprint arXiv:2512.10791},
  year={2025}
}

@article{huang2025survey,
  title={A survey on hallucination in large language models: Principles, taxonomy, challenges, and open questions},
  author={Huang, Lei and Yu, Weijiang and Ma, Weitao and Zhong, Weihong and Feng, Zhangyin and Wang, Haotian and Chen, Qianglong and Peng, Weihua and Feng, Xiaocheng and Qin, Bing and others},
  journal={ACM Transactions on Information Systems},
  volume={43},
  number={2},
  pages={1--55},
  year={2025},
  publisher={ACM New York, NY}
}

@article{zhang2025siren,
  title={Siren’s Song in the AI Ocean: A Survey on Hallucination in Large Language Models},
  author={Zhang, Yue and Li, Yafu and Cui, Leyang and Cai, Deng and Liu, Lemao and Fu, Tingchen and Huang, Xinting and Zhao, Enbo and Zhang, Yu and Chen, Yulong and others},
  journal={Computational Linguistics},
  pages={1--46},
  year={2025},
  publisher={MIT Press 255 Main Street, 9th Floor, Cambridge, Massachusetts 02142, USA~…}
}

@misc{jackson2025aaomniscienceevaluatingcrossdomainknowledge,
      title={AA-Omniscience: Evaluating Cross-Domain Knowledge Reliability in Large Language Models}, 
      author={Declan Jackson and William Keating and George Cameron and Micah Hill-Smith},
      year={2025},
      eprint={2511.13029},
      archivePrefix={arXiv},
      primaryClass={cs.CL},
      url={https://arxiv.org/abs/2511.13029}, 
}

@article{ji2023survey,
  title={Survey of hallucination in natural language generation},
  author={Ji, Ziwei and Lee, Nayeon and Frieske, Rita and Yu, Tiezheng and Su, Dan and Xu, Yan and Ishii, Etsuko and Bang, Ye Jin and Madotto, Andrea and Fung, Pascale},
  journal={ACM Computing Surveys},
  volume={55},
  number={12},
  pages={1--38},
  year={2023},
  publisher={ACM New York, NY}
}

@inproceedings{kandpal2023large,
  title={Large language models struggle to learn long-tail knowledge},
  author={Kandpal, Nikhil and Deng, Haikang and Roberts, Adam and Wallace, Eric and Raffel, Colin},
  booktitle={International Conference on Machine Learning},
  pages={15696--15707},
  year={2023},
  organization={PMLR}
}

@inproceedings{mallen2023not,
  title={When not to trust language models: Investigating effectiveness of parametric and non-parametric memories},
  author={Mallen, Alex and Asai, Akari and Zhong, Victor and Das, Rajarshi and Khashabi, Daniel and Hajishirzi, Hannaneh},
  booktitle={Proceedings of the 61st Annual Meeting of the Association for Computational Linguistics (Volume 1: Long Papers)},
  pages={9802--9822},
  year={2023}
}

@article{liu2025metafaith,
  title={MetaFaith: Faithful Natural Language Uncertainty Expression in LLMs},
  author={Liu, Gabrielle Kaili-May and Yona, Gal and Caciularu, Avi and Szpektor, Idan and Rudner, Tim GJ and Cohan, Arman},
  journal={arXiv preprint arXiv:2505.24858},
  year={2025}
}

@inproceedings{yona2024can,
  title={Can Large Language Models Faithfully Express Their Intrinsic Uncertainty in Words?},
  author={Yona, Gal and Aharoni, Roee and Geva, Mor},
  booktitle={Proceedings of the 2024 Conference on Empirical Methods in Natural Language Processing},
  pages={7752--7764},
  year={2024}
}

@article{haas2025simpleqa,
  title={SimpleQA Verified: A reliable factuality benchmark to measure parametric knowledge},
  author={Haas, Lukas and Yona, Gal and D'Antonio, Giovanni and Goldshtein, Sasha and Das, Dipanjan},
  journal={arXiv preprint arXiv:2509.07968},
  year={2025}
}

@article{schick2023toolformer,
  title={Toolformer: Language models can teach themselves to use tools},
  author={Schick, Timo and Dwivedi-Yu, Jane and Dess{\`\i}, Roberto and Raileanu, Roberta and Lomeli, Maria and Hambro, Eric and Zettlemoyer, Luke and Cancedda, Nicola and Scialom, Thomas},
  journal={Advances in Neural Information Processing Systems},
  volume={36},
  pages={68539--68551},
  year={2023}
}

@article{lin2021truthfulqa,
  title={Truthfulqa: Measuring how models mimic human falsehoods},
  author={Lin, Stephanie and Hilton, Jacob and Evans, Owain},
  journal={arXiv preprint arXiv:2109.07958},
  year={2021}
}

@article{wei2024measuring,
  title={Measuring short-form factuality in large language models},
  author={Wei, Jason and Karina, Nguyen and Chung, Hyung Won and Jiao, Yunxin Joy and Papay, Spencer and Glaese, Amelia and Schulman, John and Fedus, William},
  journal={arXiv preprint arXiv:2411.04368},
  year={2024}
}

@article{kadavath2022language,
  title={Language models (mostly) know what they know},
  author={Kadavath, Saurav and Conerly, Tom and Askell, Amanda and Henighan, Tom and Drain, Dawn and Perez, Ethan and Schiefer, Nicholas and Hatfield-Dodds, Zac and DasSarma, Nova and Tran-Johnson, Eli and others},
  journal={arXiv preprint arXiv:2207.05221},
  year={2022}
}

@inproceedings{petroni2019language,
  title={Language models as knowledge bases?},
  author={Petroni, Fabio and Rockt{\"a}schel, Tim and Riedel, Sebastian and Lewis, Patrick and Bakhtin, Anton and Wu, Yuxiang and Miller, Alexander},
  booktitle={Proceedings of the 2019 conference on empirical methods in natural language processing and the 9th international joint conference on natural language processing (EMNLP-IJCNLP)},
  pages={2463--2473},
  year={2019}
}

@inproceedings{roberts2020much,
  title={How Much Knowledge Can You Pack Into the Parameters of a Language Model?},
  author={Roberts, Adam and Raffel, Colin and Shazeer, Noam},
  booktitle={Proceedings of the 2020 Conference on Empirical Methods in Natural Language Processing (EMNLP)},
  pages={5418--5426},
  year={2020}
}

@article{lewis2020retrieval,
  title={Retrieval-augmented generation for knowledge-intensive nlp tasks},
  author={Lewis, Patrick and Perez, Ethan and Piktus, Aleksandra and Petroni, Fabio and Karpukhin, Vladimir and Goyal, Naman and K{\"u}ttler, Heinrich and Lewis, Mike and Yih, Wen-tau and Rockt{\"a}schel, Tim and others},
  journal={Advances in neural information processing systems},
  volume={33},
  pages={9459--9474},
  year={2020}
}

@inproceedings{yao2022react,
  title={React: Synergizing reasoning and acting in language models},
  author={Yao, Shunyu and Zhao, Jeffrey and Yu, Dian and Du, Nan and Shafran, Izhak and Narasimhan, Karthik R and Cao, Yuan},
  booktitle={The eleventh international conference on learning representations},
  year={2022}
}

@article{nakano2021webgpt,
  title={Webgpt: Browser-assisted question-answering with human feedback},
  author={Nakano, Reiichiro and Hilton, Jacob and Balaji, Suchir and Wu, Jeff and Ouyang, Long and Kim, Christina and Hesse, Christopher and Jain, Shantanu and Kosaraju, Vineet and Saunders, William and others},
  journal={arXiv preprint arXiv:2112.09332},
  year={2021}
}

@article{ouyang2022training,
  title={Training language models to follow instructions with human feedback},
  author={Ouyang, Long and Wu, Jeffrey and Jiang, Xu and Almeida, Diogo and Wainwright, Carroll and Mishkin, Pamela and Zhang, Chong and Agarwal, Sandhini and Slama, Katarina and Ray, Alex and others},
  journal={Advances in neural information processing systems},
  volume={35},
  pages={27730--27744},
  year={2022}
}

@inproceedings{tian2023fine,
  title={Fine-tuning language models for factuality},
  author={Tian, Katherine and Mitchell, Eric and Yao, Huaxiu and Manning, Christopher D and Finn, Chelsea},
  booktitle={The Twelfth International Conference on Learning Representations},
  year={2023}
}

@inproceedings{chuangdola,
  title={DoLa: Decoding by Contrasting Layers Improves Factuality in Large Language Models},
  author={Chuang, Yung-Sung and Xie, Yujia and Luo, Hongyin and Kim, Yoon and Glass, James R and He, Pengcheng},
  booktitle={The Twelfth International Conference on Learning Representations}
}

@article{xu2024hallucination,
  title={Hallucination is inevitable: An innate limitation of large language models},
  author={Xu, Ziwei and Jain, Sanjay and Kankanhalli, Mohan},
  journal={arXiv preprint arXiv:2401.11817},
  year={2024}
}

@inproceedings{kalavasis2025limits,
  title={On the Limits of Language Generation: Trade-Offs between Hallucination and Mode-Collapse},
  author={Kalavasis, Alkis and Mehrotra, Anay and Velegkas, Grigoris},
  booktitle={Proceedings of the 57th Annual ACM Symposium on Theory of Computing},
  pages={1732--1743},
  year={2025}
}

@inproceedings{banerjee2025llms,
  title={Llms will always hallucinate, and we need to live with this},
  author={Banerjee, Sourav and Agarwal, Ayushi and Singla, Saloni},
  booktitle={Intelligent Systems Conference},
  pages={624--648},
  year={2025},
  organization={Springer}
}

@article{tao2025revisiting,
  title={Revisiting Uncertainty Estimation and Calibration of Large Language Models},
  author={Tao, Linwei and Yeh, Yi-Fan and Dong, Minjing and Huang, Tao and Torr, Philip and Xu, Chang},
  journal={arXiv preprint arXiv:2505.23854},
  year={2025}
}

@article{ghafouri2024epistemic,
  title={Epistemic Integrity in Large Language Models},
  author={Ghafouri, Bijean and Mohammadzadeh, Shahrad and Zhou, James and Nair, Pratheeksha and Tian, Jacob-Junqi and Tsujimura, Hikaru and Goel, Mayank and Krishna, Sukanya and Rabbany, Reihaneh and Godbout, Jean-Fran{\c{c}}ois and others},
  journal={arXiv preprint arXiv:2411.06528},
  year={2024}
}

@article{wang2025bias,
  title={When Bias Pretends to Be Truth: How Spurious Correlations Undermine Hallucination Detection in LLMs},
  author={Wang, Shaowen and Dong, Yiqi and Chang, Ruinian and Zhu, Tansheng and Sun, Yuebo and Lyu, Kaifeng and Li, Jian},
  journal={arXiv preprint arXiv:2511.07318},
  year={2025}
}

@article{simhi2025hack,
  title={HACK: Hallucinations Along Certainty and Knowledge Axes},
  author={Simhi, Adi and Herzig, Jonathan and Itzhak, Itay and Arad, Dana and Gekhman, Zorik and Reichart, Roi and Barez, Fazl and Stanovsky, Gabriel and Szpektor, Idan and Belinkov, Yonatan},
  journal={arXiv preprint arXiv:2510.24222},
  year={2025}
}

@article{joglekar2025training,
  title={Training LLMs for Honesty via Confessions},
  author={Joglekar, Manas and Chen, Jeremy and Wu, Gabriel and Yosinski, Jason and Wang, Jasmine and Barak, Boaz and Glaese, Amelia},
  journal={arXiv preprint arXiv:2512.08093},
  year={2025}
}

@article{yao2025reasoning,
  title={Are Reasoning Models More Prone to Hallucination?},
  author={Yao, Zijun and Liu, Yantao and Chen, Yanxu and Chen, Jianhui and Fang, Junfeng and Hou, Lei and Li, Juanzi and Chua, Tat-Seng},
  journal={arXiv preprint arXiv:2505.23646},
  year={2025}
}

@article{jaech2024openai,
  title={Openai o1 system card},
  author={Jaech, Aaron and Kalai, Adam and Lerer, Adam and Richardson, Adam and El-Kishky, Ahmed and Low, Aiden and Helyar, Alec and Madry, Aleksander and Beutel, Alex and Carney, Alex and others},
  journal={arXiv preprint arXiv:2412.16720},
  year={2024}
}

@inproceedings{marks2023geometry,
  title={The Geometry of Truth: Emergent Linear Structure in Large Language Model Representations of True/False Datasets},
  author={Marks, Samuel and Tegmark, Max},
  booktitle={First Conference on Language Modeling}
}

@inproceedings{sky2024androids,
  title={Do androids know they’re only dreaming of electric sheep?},
  author={Sky, CH-Wang and Van Durme, Benjamin and Eisner, Jason and Kedzie, Chris},
  booktitle={Findings of the Association for Computational Linguistics: ACL 2024},
  pages={4401--4420},
  year={2024}
}

@inproceedings{orgad2024llms,
  title={LLMs Know More Than They Show: On the Intrinsic Representation of LLM Hallucinations},
  author={Orgad, Hadas and Toker, Michael and Gekhman, Zorik and Reichart, Roi and Szpektor, Idan and Kotek, Hadas and Belinkov, Yonatan},
  booktitle={ICLR},
  year={2025}
}

@inproceedings{joshi2017triviaqa,
  title={TriviaQA: A Large Scale Distantly Supervised Challenge Dataset for Reading Comprehension},
  author={Joshi, Mandar and Choi, Eunsol and Weld, Daniel S and Zettlemoyer, Luke},
  booktitle={Proceedings of the 55th Annual Meeting of the Association for Computational Linguistics (Volume 1: Long Papers)},
  pages={1601--1611},
  year={2017}
}

@inproceedings{binderlooking,
  title={Looking Inward: Language Models Can Learn About Themselves by Introspection},
  author={Binder, Felix Jedidja and Chua, James and Korbak, Tomek and Sleight, Henry and Hughes, John and Long, Robert and Perez, Ethan and Turpin, Miles and Evans, Owain},
  booktitle={The Thirteenth International Conference on Learning Representations}
}

@inproceedings{liu2023cognitive,
  title={Cognitive Dissonance: Why Do Language Model Outputs Disagree with Internal Representations of Truthfulness?},
  author={Liu, Kevin and Casper, Stephen and Hadfield-Menell, Dylan and Andreas, Jacob},
  booktitle={Proceedings of the 2023 Conference on Empirical Methods in Natural Language Processing},
  pages={4791--4797},
  year={2023}
}

@article{kirichenko2025abstentionbench,
  title={AbstentionBench: Reasoning LLMs Fail on Unanswerable Questions},
  author={Kirichenko, Polina and Ibrahim, Mark and Chaudhuri, Kamalika and Bell, Samuel J},
  journal={arXiv preprint arXiv:2506.09038},
  year={2025}
}

@article{zhang2025factguard,
  title={FactGuard: Leveraging Multi-Agent Systems to Generate Answerable and Unanswerable Questions for Enhanced Long-Context LLM Extraction},
  author={Zhang, Qian-Wen and Li, Fang and Wang, Jie and Qiao, Lingfeng and Yu, Yifei and Yin, Di and Sun, Xing},
  journal={arXiv preprint arXiv:2504.05607},
  year={2025}
}

@article{kapoor2024large,
  title={Large language models must be taught to know what they don’t know},
  author={Kapoor, Sanyam and Gruver, Nate and Roberts, Manley and Collins, Katie and Pal, Arka and Bhatt, Umang and Weller, Adrian and Dooley, Samuel and Goldblum, Micah and Wilson, Andrew G},
  journal={Advances in Neural Information Processing Systems},
  volume={37},
  pages={85932--85972},
  year={2024}
}

@article{gekhman2025inside,
  title={Inside-out: Hidden factual knowledge in llms},
  author={Gekhman, Zorik and David, Eyal Ben and Orgad, Hadas and Ofek, Eran and Belinkov, Yonatan and Szpektor, Idan and Herzig, Jonathan and Reichart, Roi},
  journal={arXiv preprint arXiv:2503.15299},
  year={2025}
}

@article{ravfogel2025emergence,
  title={Emergence of linear truth encodings in language models},
  author={Ravfogel, Shauli and Yehudai, Gilad and Linzen, Tal and Bruna, Joan and Bietti, Alberto},
  journal={arXiv preprint arXiv:2510.15804},
  year={2025}
}

@article{li2025training,
  title={Training Language Models to Explain Their Own Computations},
  author={Li, Belinda Z and Guo, Zifan Carl and Huang, Vincent and Steinhardt, Jacob and Andreas, Jacob},
  journal={arXiv preprint arXiv:2511.08579},
  year={2025}
}

@article{lindsey2025emergent,
  author={Lindsey, Jack},
  title={Emergent Introspective Awareness in Large Language Models},
  journal={Transformer Circuits Thread},
  year={2025},
  url={https://transformer-circuits.pub/2025/introspection/index.html}
}

@article{delacroix2025beyond,
  title={Beyond quantification: Navigating uncertainty in professional AI systems},
  author={Delacroix, Sylvie and Robinson, Diana and Bhatt, Umang and Domenicucci, Jacopo and Montgomery, Jessica and Varoquaux, Ga{\"e}l and Ek, Carl Henrik and Fortuin, Vincent and He, Yulan and Diethe, Tom and others},
  journal={RSS: Data Science and Artificial Intelligence},
  volume={1},
  number={1},
  pages={udaf002},
  year={2025},
  publisher={Oxford University Press UK}
}

@article{song2025outcome,
  title={Outcome-based exploration for llm reasoning},
  author={Song, Yuda and Kempe, Julia and Munos, Remi},
  journal={arXiv preprint arXiv:2509.06941},
  year={2025}
}

@inproceedings{he2025rewarding,
  title={Rewarding the unlikely: Lifting grpo beyond distribution sharpening},
  author={He, Andre Wang and Fried, Daniel and Welleck, Sean},
  booktitle={Proceedings of the 2025 Conference on Empirical Methods in Natural Language Processing},
  pages={25559--25571},
  year={2025}
}

@inproceedings{zhu2023calibration,
  title={On the Calibration of Large Language Models and Alignment},
  author={Zhu, Chiwei and Xu, Benfeng and Wang, Quan and Zhang, Yongdong and Mao, Zhendong},
  booktitle={Findings of the Association for Computational Linguistics: EMNLP 2023},
  pages={9778--9795},
  year={2023}
}

@article{achiam2023gpt,
  title={Gpt-4 technical report},
  author={Achiam, Josh and Adler, Steven and Agarwal, Sandhini and Ahmad, Lama and Akkaya, Ilge and Aleman, Florencia Leoni and Almeida, Diogo and Altenschmidt, Janko and Altman, Sam and Anadkat, Shyamal and others},
  journal={arXiv preprint arXiv:2303.08774},
  year={2023}
}

@inproceedings{yona2025keep,
  title={Keep Guessing? When Considering Inference Scaling, Mind the Baselines},
  author={Yona, Gal and Honovich, Or and Levy, Omer and Aharoni, Roee},
  booktitle={Findings of the Association for Computational Linguistics: NAACL 2025},
  pages={5979--5991},
  year={2025}
}

@article{son2002relation,
  title={The relation between metacognitive monitoring and control},
  author={Son, Lisa K and Schwartz, Bennett L},
  journal={Applied metacognition},
  pages={15--38},
  year={2002}
}

@article{james1890principles,
  title={The principles of psychology},
  author={James, William},
  journal={Henry Holt},
  year={1890}
}

@article{ji2025calibrating,
  title={Calibrating Verbal Uncertainty as a Linear Feature to Reduce Hallucinations},
  author={Ji, Ziwei and Yu, Lei and Koishekenov, Yeskendir and Bang, Yejin and Hartshorn, Anthony and Schelten, Alan and Zhang, Cheng and Fung, Pascale and Cancedda, Nicola},
  journal={arXiv preprint arXiv:2503.14477},
  year={2025}
}

@inproceedings{zhou2024relying,
  title={Relying on the Unreliable: The Impact of Language Models’ Reluctance to Express Uncertainty},
  author={Zhou, Kaitlyn and Hwang, Jena and Ren, Xiang and Sap, Maarten},
  booktitle={Proceedings of the 62nd Annual Meeting of the Association for Computational Linguistics (Volume 1: Long Papers)},
  pages={3623--3643},
  year={2024}
}

@inproceedings{kim2024m,
  title={" I'm Not Sure, But...": Examining the Impact of Large Language Models' Uncertainty Expression on User Reliance and Trust},
  author={Kim, Sunnie SY and Liao, Q Vera and Vorvoreanu, Mihaela and Ballard, Stephanie and Vaughan, Jennifer Wortman},
  booktitle={Proceedings of the 2024 ACM conference on fairness, accountability, and transparency},
  pages={822--835},
  year={2024}
}

@inproceedings{blasioksmooth,
  title={Smooth ECE: Principled Reliability Diagrams via Kernel Smoothing},
  author={Blasiok, Jaroslaw and Nakkiran, Preetum},
  booktitle={The Twelfth International Conference on Learning Representations}
}

@inproceedings{yu2024mechanistic,
  title={Mechanistic understanding and mitigation of language model non-factual hallucinations},
  author={Yu, Lei and Cao, Meng and Cheung, Jackie CK and Dong, Yue},
  booktitle={Findings of the Association for Computational Linguistics: EMNLP 2024},
  pages={7943--7956},
  year={2024}
}

@article{steyvers2025large,
  title={What large language models know and what people think they know},
  author={Steyvers, Mark and Tejeda, Heliodoro and Kumar, Aakriti and Belem, Catarina and Karny, Sheer and Hu, Xinyue and Mayer, Lukas W and Smyth, Padhraic},
  journal={Nature Machine Intelligence},
  volume={7},
  number={2},
  pages={221--231},
  year={2025},
  publisher={Nature Publishing Group UK London}
}

@article{levinstein2023still,
  title={Still no lie detector for language models: Probing empirical and conceptual roadblocks},
  author={Levinstein, Benjamin A and Herrmann, Daniel A},
  journal={arXiv preprint arXiv:2307.00175},
  year={2023}
}

@article{li2025hallucination,
  title={The Hallucination Dilemma: Factuality-Aware Reinforcement Learning for Large Reasoning Models},
  author={Li, Junyi and Ng, Hwee Tou},
  journal={arXiv preprint arXiv:2505.24630},
  year={2025}
}

@article{rabanser2026towards,
  title={Towards a science of AI agent reliability},
  author={Rabanser, Stephan and Kapoor, Sayash and Kirgis, Peter and Liu, Kangheng and Utpala, Saiteja and Narayanan, Arvind},
  journal={arXiv preprint arXiv:2602.16666},
  year={2026}
}

@article{kaplan2026fine,
  title={Why Fine-Tuning Encourages Hallucinations and How to Fix It},
  author={Kaplan, Guy and Gekhman, Zorik and Zhu, Zhen and Rozner, Lotem and Reif, Yuval and Swayamdipta, Swabha and Hoiem, Derek and Schwartz, Roy},
  journal={arXiv preprint arXiv:2604.15574},
  year={2026}
}

@article{xu2026trust,
  title={When to Trust Tools? Adaptive Tool Trust Calibration For Tool-Integrated Math Reasoning},
  author={Xu, Ruotao and Ji, Yixin and Luo, Yu and Li, Jinpeng and Li, Dong and Li, Peifeng and Li, Juntao and Zhang, Min},
  journal={arXiv preprint arXiv:2604.08281},
  year={2026}
}

@article{yan2026act,
  title={Act Wisely: Cultivating Meta-Cognitive Tool Use in Agentic Multimodal Models},
  author={Yan, Shilin and Tong, Jintao and Xue, Hongwei and Tang, Xiaojun and Wang, Yangyang and Shi, Kunyu and Zhang, Guannan and Li, Ruixuan and Zou, Yixiong},
  journal={arXiv preprint arXiv:2604.08545},
  year={2026}
}

@article{lin2022teaching,
  title={Teaching models to express their uncertainty in words},
  author={Lin, Stephanie and Hilton, Jacob and Evans, Owain},
  journal={arXiv preprint arXiv:2205.14334},
  year={2022}
}

@article{savage2025large,
  title={Large language model uncertainty proxies: discrimination and calibration for medical diagnosis and treatment},
  author={Savage, Thomas and Wang, John and Gallo, Robert and Boukil, Abdessalem and Patel, Vishwesh and Safavi-Naini, Seyed Amir Ahmad and Soroush, Ali and Chen, Jonathan H},
  journal={Journal of the American Medical Informatics Association},
  volume={32},
  number={1},
  pages={139--149},
  year={2025},
  publisher={Oxford University Press}
}

@article{nakkiran2025trained,
  title={Trained on Tokens, Calibrated on Concepts: The Emergence of Semantic Calibration in LLMs},
  author={Nakkiran, Preetum and Bradley, Arwen and Goli{\'n}ski, Adam and Ndiaye, Eugene and Kirchhof, Michael and Williamson, Sinead},
  journal={arXiv preprint arXiv:2511.04869},
  year={2025}
}

@inproceedings{tian2023just,
  title={Just Ask for Calibration: Strategies for Eliciting Calibrated Confidence Scores from Language Models Fine-Tuned with Human Feedback},
  author={Tian, Katherine and Mitchell, Eric and Zhou, Allan and Sharma, Archit and Rafailov, Rafael and Yao, Huaxiu and Finn, Chelsea and Manning, Christopher D},
  booktitle={Proceedings of the 2023 Conference on Empirical Methods in Natural Language Processing},
  pages={5433--5442},
  year={2023}
}

@inproceedings{zhao2024fact,
  title={Fact-and-Reflection (FaR) Improves Confidence Calibration of Large Language Models},
  author={Zhao, Xinran and Zhang, Hongming and Pan, Xiaoman and Yao, Wenlin and Yu, Dong and Wu, Tongshuang and Chen, Jianshu},
  booktitle={Findings of the Association for Computational Linguistics ACL 2024},
  pages={8702--8718},
  year={2024}
}

@inproceedings{gekhman2024does,
  title={Does Fine-Tuning LLMs on New Knowledge Encourage Hallucinations?},
  author={Gekhman, Zorik and Yona, Gal and Aharoni, Roee and Eyal, Matan and Feder, Amir and Reichart, Roi and Herzig, Jonathan},
  booktitle={Proceedings of the 2024 Conference on Empirical Methods in Natural Language Processing},
  pages={7765--7784},
  year={2024}
}

@article{taubenfeld2025confidence,
  title={Confidence improves self-consistency in llms},
  author={Taubenfeld, Amir and Sheffer, Tom and Ofek, Eran and Feder, Amir and Goldstein, Ariel and Gekhman, Zorik and Yona, Gal},
  journal={arXiv preprint arXiv:2502.06233},
  year={2025}
}

@inproceedings{cohen2023lm,
  title={LM vs LM: Detecting Factual Errors via Cross Examination},
  author={Cohen, Roi and Hamri, May and Geva, Mor and Globerson, Amir},
  booktitle={Proceedings of the 2023 Conference on Empirical Methods in Natural Language Processing},
  pages={12621--12640},
  year={2023}
}

@inproceedings{dhuliawala2024chain,
  title={Chain-of-verification reduces hallucination in large language models},
  author={Dhuliawala, Shehzaad and Komeili, Mojtaba and Xu, Jing and Raileanu, Roberta and Li, Xian and Celikyilmaz, Asli and Weston, Jason},
  booktitle={Findings of the association for computational linguistics: ACL 2024},
  pages={3563--3578},
  year={2024}
}

@inproceedings{kalai2024calibrated,
  title={Calibrated language models must hallucinate},
  author={Kalai, Adam Tauman and Vempala, Santosh S},
  booktitle={Proceedings of the 56th Annual ACM Symposium on Theory of Computing},
  pages={160--171},
  year={2024}
}

@article{kwiatkowski2019natural,
  title={Natural questions: a benchmark for question answering research},
  author={Kwiatkowski, Tom and Palomaki, Jennimaria and Redfield, Olivia and Collins, Michael and Parikh, Ankur and Alberti, Chris and Epstein, Danielle and Polosukhin, Illia and Devlin, Jacob and Lee, Kenton and others},
  journal={Transactions of the Association for Computational Linguistics},
  volume={7},
  pages={453--466},
  year={2019},
  publisher={MIT Press One Rogers Street, Cambridge, MA 02142-1209, USA journals-info~…}
}

@inproceedings{shi2024trusting,
  title={Trusting your evidence: Hallucinate less with context-aware decoding},
  author={Shi, Weijia and Han, Xiaochuang and Lewis, Mike and Tsvetkov, Yulia and Zettlemoyer, Luke and Yih, Wen-tau},
  booktitle={Proceedings of the 2024 Conference of the North American Chapter of the Association for Computational Linguistics: Human Language Technologies (Volume 2: Short Papers)},
  pages={783--791},
  year={2024}
}

@article{li2023inference,
  title={Inference-time intervention: Eliciting truthful answers from a language model},
  author={Li, Kenneth and Patel, Oam and Vi{\'e}gas, Fernanda and Pfister, Hanspeter and Wattenberg, Martin},
  journal={Advances in Neural Information Processing Systems},
  volume={36},
  pages={41451--41530},
  year={2023}
}

@article{yoon2025reasoning,
  title={Reasoning models better express their confidence},
  author={Yoon, Dongkeun and Kim, Seungone and Yang, Sohee and Kim, Sunkyoung and Kim, Soyeon and Kim, Yongil and Choi, Eunbi and Kim, Yireun and Seo, Minjoon},
  journal={arXiv preprint arXiv:2505.14489},
  year={2025}
}

@article{podolak2025read,
  title={Read Your Own Mind: Reasoning Helps Surface Self-Confidence Signals in LLMs},
  author={Podolak, Jakub and Verma, Rajeev},
  journal={arXiv preprint arXiv:2505.23845},
  year={2025}
}

@article{stolfo2024confidence,
  title={Confidence regulation neurons in language models},
  author={Stolfo, Alessandro and Wu, Ben and Gurnee, Wes and Belinkov, Yonatan and Song, Xingyi and Sachan, Mrinmaya and Nanda, Neel},
  journal={Advances in Neural Information Processing Systems},
  volume={37},
  pages={125019--125049},
  year={2024}
}

@article{li2025confidence,
  title={Confidence Is All You Need: Few-Shot RL Fine-Tuning of Language Models},
  author={Li, Pengyi and Skripkin, Matvey and Zubrey, Alexander and Kuznetsov, Andrey and Oseledets, Ivan},
  journal={arXiv preprint arXiv:2506.06395},
  year={2025}
}

@article{chentanez2004intrinsically,
  title={Intrinsically motivated reinforcement learning},
  author={Chentanez, Nuttapong and Barto, Andrew and Singh, Satinder},
  journal={Advances in neural information processing systems},
  volume={17},
  year={2004}
}

@inproceedings{sukhija2025optimism,
  title={Optimism via intrinsic rewards: Scalable and principled exploration for model-based reinforcement learning},
  author={Sukhija, Bhavya and Treven, Lenart and Sferrazza, Carmelo and Dorfler, Florian and Abbeel, Pieter and Krause, Andreas},
  booktitle={7th Robot Learning Workshop: Towards Robots with Human-Level Abilities},
  year={2025}
}

@article{wang2025icpo,
  title={ICPO: Intrinsic Confidence-Driven Group Relative Preference Optimization for Efficient Reinforcement Learning},
  author={Wang, Jinpeng and Li, Chao and Ye, Ting and Zhang, Mengyuan and Liu, Wei and Luan, Jian},
  journal={arXiv preprint arXiv:2511.21005},
  year={2025}
}

@article{kayal2025impact,
  title={The impact of intrinsic rewards on exploration in Reinforcement Learning},
  author={Kayal, Aya and Pignatelli, Eduardo and Toni, Laura},
  journal={Neural Computing and Applications},
  pages={1--35},
  year={2025},
  publisher={Springer}
}

@article{prabhudesai2025maximizing,
  title={Maximizing Confidence Alone Improves Reasoning},
  author={Prabhudesai, Mihir and Chen, Lili and Ippoliti, Alex and Fragkiadaki, Katerina and Liu, Hao and Pathak, Deepak},
  journal={arXiv preprint arXiv:2505.22660},
  year={2025}
}

@article{eikema2025teaching,
  title={Teaching Language Models to Faithfully Express their Uncertainty},
  author={Eikema, Bryan and Ilia, Evgenia and de Souza, Jos{\'e} GC and Zerva, Chrysoula and Aziz, Wilker},
  journal={arXiv preprint arXiv:2510.12587},
  year={2025}
}

@article{lin2025adasearch,
  title={AdaSearch: Balancing Parametric Knowledge and Search in Large Language Models via Reinforcement Learning},
  author={Lin, Tzu-Han and Chen, Wei-Lin and Li, Chen-An and Lee, Hung-yi and Chen, Yun-Nung and Meng, Yu},
  journal={arXiv preprint arXiv:2512.16883},
  year={2025}
}

@article{petroni2020context,
  title={How context affects language models' factual predictions},
  author={Petroni, Fabio and Lewis, Patrick and Piktus, Aleksandra and Rockt{\"a}schel, Tim and Wu, Yuxiang and Miller, Alexander H and Riedel, Sebastian},
  journal={arXiv preprint arXiv:2005.04611},
  year={2020}
}

@inproceedings{li2023large,
  title={Large language models with controllable working memory},
  author={Li, Daliang and Rawat, Ankit Singh and Zaheer, Manzil and Wang, Xin and Lukasik, Michal and Veit, Andreas and Yu, Felix and Kumar, Sanjiv},
  booktitle={Findings of the association for computational linguistics: ACL 2023},
  pages={1774--1793},
  year={2023}
}

@article{mielke2022reducing,
  title={Reducing conversational agents’ overconfidence through linguistic calibration},
  author={Mielke, Sabrina J and Szlam, Arthur and Dinan, Emily and Boureau, Y-Lan},
  journal={Transactions of the Association for Computational Linguistics},
  volume={10},
  pages={857--872},
  year={2022},
  publisher={MIT Press One Broadway, 12th Floor, Cambridge, Massachusetts 02142, USA~…}
}

@article{stengel2024lacie,
  title={LACIE: Listener-aware finetuning for calibration in large language models},
  author={Stengel-Eskin, Elias and Hase, Peter and Bansal, Mohit},
  journal={Advances in Neural Information Processing Systems},
  volume={37},
  pages={43080--43106},
  year={2024}
}

@article{yang2024alignment,
  title={Alignment for honesty},
  author={Yang, Yuqing and Chern, Ethan and Qiu, Xipeng and Neubig, Graham and Liu, Pengfei},
  journal={Advances in Neural Information Processing Systems},
  volume={37},
  pages={63565--63598},
  year={2024}
}

@article{eisenstein2025don,
  title={Don't lie to your friends: Learning what you know from collaborative self-play},
  author={Eisenstein, Jacob and Aghajani, Reza and Fisch, Adam and Dua, Dheeru and Huot, Fantine and Lapata, Mirella and Zayats, Vicky and Berant, Jonathan},
  journal={arXiv preprint arXiv:2503.14481},
  year={2025}
}

@article{wang2024survey,
  title={A survey on large language model based autonomous agents},
  author={Wang, Lei and Ma, Chen and Feng, Xueyang and Zhang, Zeyu and Yang, Hao and Zhang, Jingsen and Chen, Zhiyuan and Tang, Jiakai and Chen, Xu and Lin, Yankai and others},
  journal={Frontiers of Computer Science},
  volume={18},
  number={6},
  pages={186345},
  year={2024},
  publisher={Springer}
}

@inproceedings{qian2025smart,
  title={SMART: Self-aware agent for tool overuse mitigation},
  author={Qian, Cheng and Acikgoz, Emre Can and Wang, Hongru and Chen, Xiusi and Sil, Avirup and Hakkani-Tur, Dilek and Tur, Gokhan and Ji, Heng},
  booktitle={Findings of the Association for Computational Linguistics: ACL 2025},
  pages={4604--4621},
  year={2025}
}

\appendix
\section{Additional Details}
\label{sec:appendix}

\paragraph{Figure \ref{fig:calibration_discrimination} Simulation Methodology.} We created a synthetic dataset ($N=25,000$)  designed to reproduce the empirical confidence profiles reported by \citet{nakkiran2025trained}.
We fixed a base hallucination rate of 25\%. Confidence scores for correct answers ($y=1$) and incorrect answers ($y=0$) were sampled from Beta distributions, $\text{Beta}(\alpha, \beta)$, chosen to model overlapping confidence profiles typical of modern LLMs. Specifically, we sampled correct scores from $\text{Beta}(1.8, 1.0)$ (skewed toward high confidence) and incorrect scores from $\text{Beta}(1.0, 1.3)$ (skewed toward low confidence). To isolate discriminative power as the limiting factor, we applied Isotonic Regression to the raw scores. This enforced near-perfect calibration ($\text{smECE} \approx 0.014$), ensuring that the observed utility-error trade-offs stem purely from the overlapping distributions rather than miscalibrated probabilities. The utility-error trade-off curve was computed by sweeping a rejection threshold $\tau \in [0,1]$ across the calibrated scores, calculating the proportion of total samples that were answered correctly (Utility) versus answered incorrectly (Error Rate) at each threshold.

\section{Faithful Uncertainty: Definitions and Measurement}
\label{sec:appendix:faithfulness}

We provide a concise overview of the key concepts necessary to operationalize the notion of faithful uncertainty (\S\ref{sec:faithfulness}),
drawing on \citet{yona2024can}.

\paragraph{Intrinsic uncertainty.}

We quantify uncertainty via the likelihood of generating conflicting answers under repeated sampling. Specifically, 
given a fact-seeking query $Q$ (e.g., ``When was Barack
Obama born?'') and a candidate assertion $A$ (e.g. ``1961''), 
if the  assertions $A_1, \dots, A_k$ generated by $M$ under repeated sampling do not contradict $A$ (e.g., ``1961'',
``I think he was born in 1961'', or ``August 4, 1961.''),
then we say the intrinsic confidence is high of $M$ in $A$ is high: 
\begin{equation}
  \mathrm{conf}_M(A) \;\equiv\; 1 - \tfrac{1}{k}\textstyle\sum_{i=1}^{k}
  \mathbf{1}[A \text{ contradicts } A_i]
\end{equation}

\paragraph{Linguistic uncertainty.}

We quantify linguistic uncertainty via decisiveness, which reflects how confidently an assertion is
conveyed through hedges, qualifiers, and epistemic markers. Given an assertion $A$ in a model response $R$, its perceived decisiveness is captured as the probability a reader would assign to $A$ being true based solely on the language of $R$ (and is implemented using LLM-as-a-judge, see \cite{yona2024can, liu2025metafaith}):

\begin{equation}
  \mathrm{dec}(A;\,R,Q) \;=\; \Pr[A \text{ is True} \mid R, Q],
\end{equation}

\paragraph{Faithful uncertainty.}
A response $R$ \emph{faithfully expresses} $M$'s uncertainty if its
decisiveness tracks its intrinsic confidence assertion-by-assertion:
\begin{equation}
  \mathrm{faithfulness}_M(R;\,Q) \;\equiv\; 1 - \frac{1}{|A(R)|}
  \sum_{A \in A(R)} \bigl|\mathrm{dec}(A;\,R,Q) -
  \mathrm{conf}_M(A)\bigr|.
\end{equation}
A score of 1 indicates perfect alignment; lower scores reflect systematic
over- or under-hedging relative to actual internal confidence.

\paragraph{cMFG metric.}
Raw faithfulness scores are confounded by a model's confidence distribution.
\citet{yona2024can} introduce \emph{conditional mean faithful generation}
(cMFG): expected faithfulness uniformly averaged across confidence levels (in
practice, via equal-width confidence bins). A cMFG of 0.5 corresponds to a
strategy whose decisiveness is independent of actual confidence. Current
state-of-the-art models typically score 0.5--0.7, indicating that expressed
uncertainty is only weakly aligned with intrinsic confidence.

\end{document}